%% file: arxiv.tex
\documentclass[10pt,twocolumn,letterpaper]{article}

\usepackage[pagenumbers]{iccv} 
\usepackage{multirow}
\input{preamble}

\definecolor{iccvblue}{rgb}{0.21,0.49,0.74}
\usepackage[pagebackref,breaklinks,colorlinks,allcolors=iccvblue]{hyperref}

\def\paperID{11294} 
\def\confName{ICCV}
\def\confYear{2025}

\title{Beyond Training: Dynamic Token Merging for Zero-Shot Video Understanding}

\author{
\textbf{Yiming Zhang$~^{1,2}$,
Zhuokai Zhao$^{3}$,
Zhaorun Chen$^{3}$,
Zenghui Ding$^{1}$\thanks{Correspondence to Zenghui Ding: dingzenghui@iim.ac.cn.},
Xianjun Yang$^{1,2}$
Yining Sun$^{1,2}$}\\
$^{1}$HFIPS, Chinese Academy of Sciences 
$^{2}$University of Science and Technology of China \\
$^{3}$University of Chicago\\
}

\begin{document}
\maketitle
%

\input{sec/0_abstract}
\vspace{-0.2in}
\input{sec/1_intro}
\input{sec/2_related_work}
\input{sec/3_method}
\input{sec/4_experiment}
\input{sec/5_analysis}
\input{sec/6_conclusion}
\clearpage
{
    \small
    \bibliographystyle{ieeenat_fullname}
    \bibliography{main}
}

\clearpage
\input{sec/X_suppl}

\end{document}

%% file: preamble.tex
\usepackage{colortbl,xcolor}

\newcommand{\figref}[1]{Fig.~\ref{#1}}
\newcommand{\tabref}[1]{Table~\ref{#1}}
\newcommand{\eqnref}[1]{\text{Eq.}~(\ref{#1})}
\newcommand{\secref}[1]{\S\ref{#1}}

\newcommand{\algname}{\textsc{DyTo}\xspace}

\definecolor{Gray}{gray}{0.85}
\definecolor{LightCyan}{rgb}{0.88,1,1}

\newcolumntype{a}{>{\columncolor{Gray}}c}
\newcolumntype{b}{>{\columncolor{white}}c}

%% file: sec/0_abstract.tex
\begin{abstract}
Recent advancements in multimodal large language models (MLLMs) have opened new avenues for video understanding.
However, achieving high fidelity in zero-shot video tasks remains challenging.
Traditional video processing methods rely heavily on fine-tuning to capture nuanced spatial-temporal details, which incurs significant data and computation costs. 
In contrast, training-free approaches, though efficient, often lack robustness in preserving context-rich features across complex video content. 
To this end, we propose \algname, a novel dynamic token merging framework for zero-shot video understanding that adaptively optimizes token efficiency while preserving crucial scene details. 
\algname integrates a hierarchical frame selection and a bipartite token merging strategy to dynamically cluster key frames and selectively compress token sequences, striking a balance between computational efficiency with semantic richness. 
Extensive experiments across multiple benchmarks demonstrate the effectiveness of \algname, achieving superior performance compared to both fine-tuned and training-free methods and setting a new state-of-the-art for zero-shot video understanding.
Code is available at \url{https://github.com/Jam1ezhang/DYTO}.
\end{abstract}

%% file: sec/1_intro.tex
\section{Introduction}\label{sec:intro}
In recent years, video understanding has seen substantial progress, largely thanks to the rapid advancements in multimodal large language models (MLLMs)~\citep{tang2023video, zhou2024survey, madan2024foundation}. 
Traditional video understanding methods often rely on specific training to align video frames with natural language, using spatial-temporal cues to construct coherent narratives across video sequences~\citep{lin2019tsm, bertasius2021space, liu2022video}.
In contrast, MLLM-based approaches provide a more flexible and generalized framework, incorporating diverse open-world knowledge across multiple data modalities in the pre-training phase~\citep{maaz2023video, mckinzie2024mm1}. 
By utilizing this pre-trained knowledge, MLLMs can dynamically adapt to various tasks such as captioning~\citep{zhang2024ferret}, question answering~\citep{kamalloo2023evaluating}, retrieval~\citep{jiang2024vlm2vec}, and zero-shot or few-shot reasoning over various video content~\citep{chen2024fewer}. 
MLLM-based video understanding approaches generally fall into two categories, where one requires domain-specific finetuning~\citep{MbzuaioryxVideoChatGPT2024, liLLaMAVIDImageWorth2023, maVistaLLaMAReliableVideo2023, linVideoLLaVALearningUnited2023, zhao2024multimodal, zhang2024rankclip, liVideoChatChatCentricVideo2024, zhangVideoLLaMAInstructiontunedAudioVisual2023, chengVideoLLaMAAdvancingSpatialTemporal2024, jinChatUniViUnifiedVisual2024, yu2024self}, and the other is completely training-free~\citep{FreeVA, kimImageGridCan2024, chen2024halc, xuSlowFastLLaVAStrongTrainingFree2024}.

\begin{figure*}[t]
    \centering
    \includegraphics[width=\linewidth]{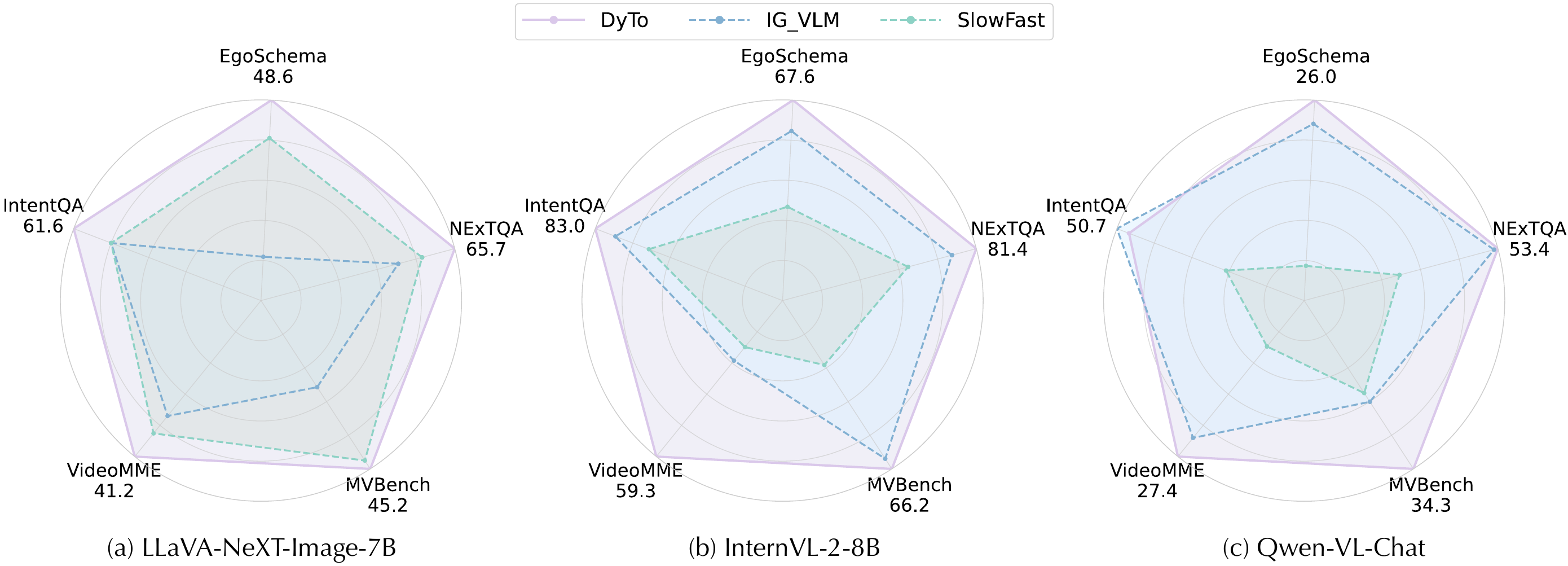}
    \caption{
        Comparison with two SOTA training-free video-based LLM decoding methods over three different model backbones across five video benchmarks. \algname and other baselines are marked using solid (—) and dashed (- - -) lines, respectively. \algname outperforms existing training-free SOTA methods on almost all the benchmarks and achieves even better performance than most SFT-based methods.
    }
    \label{fig:classification}
    \vspace{-0.15in}
\end{figure*}

Often times, trade-offs arise between accuracy and efficiency in these approaches. 
While finetuned models typically achieve higher accuracy by tailoring their capabilities to specific tasks, this comes at the cost of extensive labeled data and increased computational resources. 
In contrast, training-free methods are more efficient and flexible, leveraging generalization capabilities of pre-trained MLLMs for zero-shot inference. 
However, they may fall short in some settings, where specific reasoning tasks or varied video types require more nuanced understanding.
Thus, balancing these trade-offs remains a pivotal challenge in advancing MLLM-based video understanding.

While training-free methods offer the significant advantage of zero-shot adaptability, a central challenge lies in retaining crucial spatial-temporal details across frames without specific finetuning or frame-by-frame annotation.
Existing work has often encountered trade-offs between retaining semantic richness and maintaining computational efficiency. 
For instance, many approaches reduce frame counts and perform aggressive token pooling~\citep{FreeVA, kimImageGridCan2024, xuSlowFastLLaVAStrongTrainingFree2024}, which risk losing contextually significant frames or visual tokens that capture nuanced actions. 
Consequently, existing methods struggle to balance semantic fidelity with token efficiency in a manner that adapts to varying temporal complexities across video content.

To this end, we propose \textbf{\algname}, a novel \textbf{\textsc{dy}}namic \textbf{\textsc{to}}ken merging approach that leverages hierarchical frame selection and a fine-grained bipartite token compression method. 
\algname effectively preserves essential spatial-temporal information while significantly reducing token redundancy. 
Specifically, our method dynamically clusters key frames across hierarchical temporal scales, allowing it to adaptively capture critical events without rigid sampling strategies. 
Furthermore, by implementing a bipartite token merging mechanism, we optimize token counts based on the semantic content of each frame, which supports richer video representations under computational constraints.

By employing this dynamic, adaptive framework, \algname achieves superior performance in zero-shot video understanding, outperforming both finetuned models and other training-free methods. 
This paper’s contributions include:
\begin{itemize}
    \item A novel hierarchical bipartite merging strategy that dynamically selects key frames and performs adaptive token merging to optimize spatial-temporal fidelity and achieve more fine-grained feature retention across extended frame sequences.
    \item Comprehensive evaluation across multiple benchmarks, demonstrating \algname's superior performance in both understanding capabilities and computational efficiency over existing methods, including both finetuned and training-free approaches.
\end{itemize}

%% file: sec/2_related_work.tex
\section{Related Work}\label{sec:related_work}
\subsection{Video Large Language Models}
Video language model has witnessed significant advancements over the past years~\citep{linVideoLLaVALearningUnited2023,chengVideoLLaMAAdvancingSpatialTemporal2024,zhang2024llavanextvideo,songMovieChatDenseToken2024,xuPLLaVAParameterfreeLLaVA2024}.
Video-ChatGPT~\citep{MbzuaioryxVideoChatGPT2024} retrieves features from each frame and subsequently combines them using two operations for spatial and temporal pooling prior to feeding them into a large language model. 
LLaMA-VID~\citep{liLLaMAVIDImageWorth2023} utilized a dual token approach to effectively compress the video token by differentiating between context and content. 
Vista-LLaMA~\citep{maVistaLLaMAReliableVideo2023} presented EDVT-Attention along with a sequential vision projector that emphasizes visual tokens while decreasing temporal tokens by sequentially merging them using a Q-former. 
Video-LLaVA~\citep{linVideoLLaVALearningUnited2023} aligns the encoders for images and videos beforehand, enabling shared projections and joint training across both image and video tasks, thus mapping them into the language space. 
VideoChat~\citep{liVideoChatChatCentricVideo2024} utilized cross-attention to condense video tokens alongside user inquiries and conversational context. 
Video-LLaMA~\citep{zhangVideoLLaMAInstructiontunedAudioVisual2023} introduces a Video Q-Former and an Audio Q-Former, allowing for the integration of multiple modalities in video understanding.
In contrast, Video-LLaMA2~\citep{chengVideoLLaMAAdvancingSpatialTemporal2024} designs a spatial-temporal convolution connector to replace the Q-Former for spatial-temporal representation learning. 
Chat-UniVi~\citep{jinChatUniViUnifiedVisual2024} developed a unified model for images and videos that uses dynamic token merging with k-NN to simplify spatial and temporal tokens.  
SeViLA~\citep{yu2024self} focused on extracting key frames that are relevant to the inquiries and analyzed the video by transforming these keyframes into video tokens.

\subsection{Training-free Video LLMs}
Recent research explored and demonstrated that Image LLMs require no additional fine-tuning to apply for video understanding scenarios. 
FreeVA~\citep{FreeVA} explores the different spatial-temporal pooling strategies and versions of close-sourced GPT evaluation that influence the video understanding performance. 
IG-VLM~\citep{kimImageGridCan2024} design image grid format and assemble multiple video frames as an image before sending them to an Image LLM. 
SlowFast-LLaVA~\citep{xuSlowFastLLaVAStrongTrainingFree2024} (SF-LLaVA for short) introduced a new fusion technique for short-long sampling and various pooling strategies. 
These methods demonstrate promising results across various video benchmarks, but they have two main limitations. 
Initially, they all sampled video frames uniformly to a fixed length as the representation of the video. 
This approach inevitably loses important event information within the videos.
Although SF-LLaVA samples longer sequences than other methods, its design of using a small number of tokens fails to capture abundant spatial information present in each frame. 
Furthermore, the average or maximum pooling method employed does not adequately preserve the significant changes in action over the temporal dimension. 
In this paper, we present a new method to dynamically select frames and merge visual tokens, enabling us to comprehensively and efficiently capture the complete semantic information of every video. 
We also expand the sample video frame sequence to 100 or more frames to enhance performance on longer video understanding task.

%% file: sec/3_method.tex
\section{Method}\label{sec:method}
\begin{figure*}[t]
    \centering
    \includegraphics[width=\textwidth]{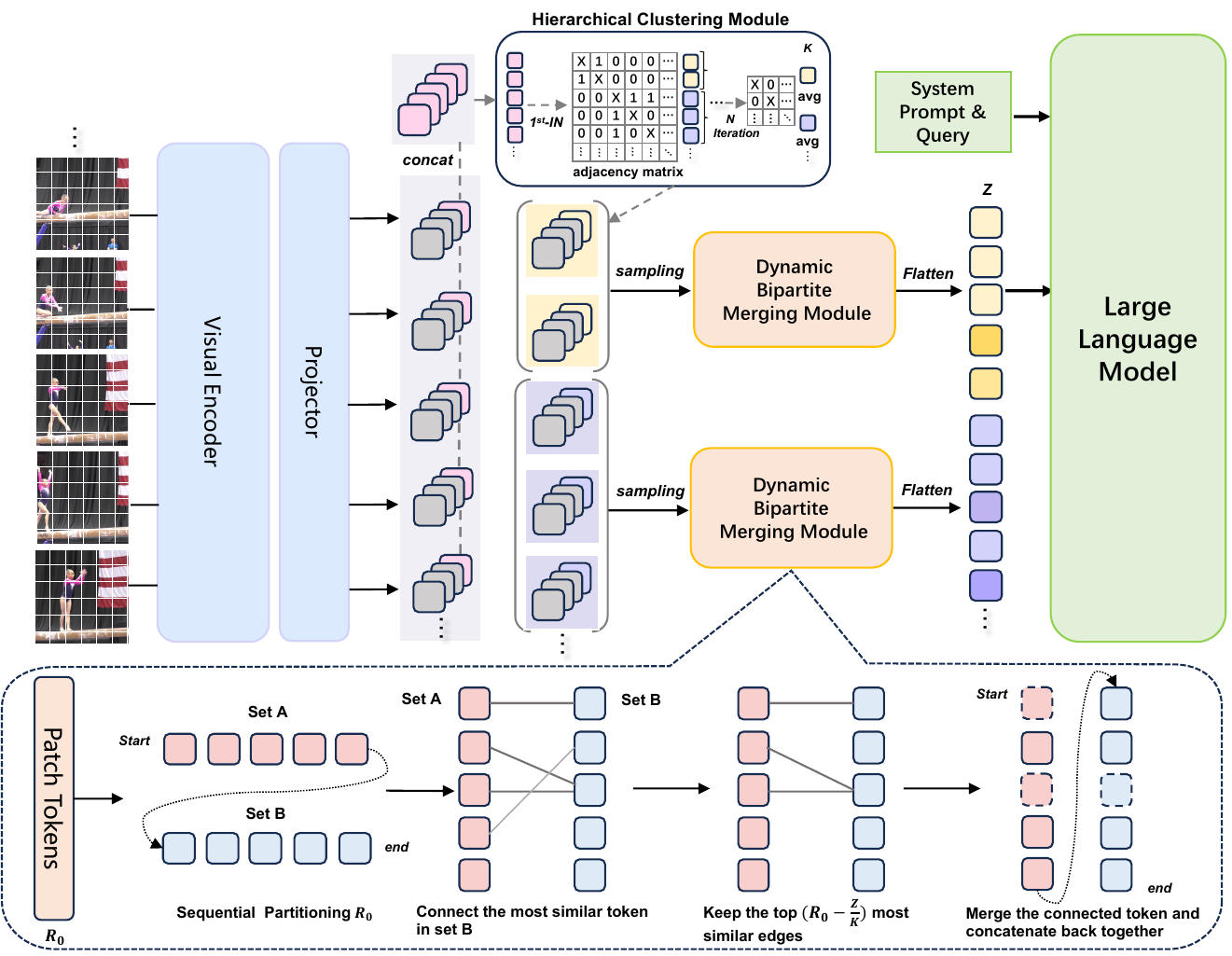}
    \caption{
        The overview of \algname, a training-free model built upon image-based MLLM without any fine-tuning. Specifically, \algname first divides the video into $K$ clusters using the [CLS] token (\textcolor{pink}{pink} block). Then the dynamic bipartite merging module samples frames from each cluster and controls the final output length as $Z$, resulting in better balance between computational efficiency and semantic richness.
    }
    \label{fig:rankclip}
    \vspace{-0.1in}
\end{figure*}
To reduce the loss of crucial information during token compression, we introduce a hierarchical bipartite graph merging mechanism. Initially, For a video that uniformly sampling with N Frames $X=\{I_1,I_2,I_3,...,I_N\}$, the frames of the video are encoded separately by visual encoders $f:Visual_{enc}(I_i)$, producing visual tokens $V \in \mathbb{R}^{ N \times L \times D}$. These are subsequently processed by the token aggregator module, transitioning from coarse-grained to fine-grained feature.

\subsection{Coarse-grained Hierarchical Clustering}~\label{subsec:hierarchical_clustering}
For $N$ frames tokens sequence $V=\{v_1,v_2,v_3,...,v_N\}$ of a video, $v_i \in \mathbb{R}^{ L \times D}$ ,we define a directed graph $G=(C,E)$ where C is the set of nodes(the points to be clustered) and the edges E connect the nodes in the feature space. 
Previous studies demonstrate the deep layers tend to aggregate global semantics in the image ~\citep{li2022uniformer,pan2022less}, therefore we collect the [CLS] tokens  of each $v_i$ and concat them as the compressed representation $\mathbf{v}\in \mathbb{R}^ {N \times D} $of the video. The time-stamps are defined as $T={1,2,...,N}$.  We build $W_t$ by calculating the feature space distances between frames and adjusting them based on their temporal positions.
\begin{equation}
W_t(i, j) = \begin{cases} 
            (1 - \langle \mathbf{v}_i, \mathbf{v}_j \rangle) \cdot {|t_i - t_j|}/{N} & \text{if } i \neq j \\
            1 & \text{otherwise}
            \end{cases}
\end{equation}
$v_i$ and $v_j$ indicate the [CLS] token embedding. The inner product is calculated on L2-normalized feature vectors to keep the distance within the [0, 1] range.
The term ${|t_i - t_j|}/{N}$ serves as a weighting mechanism based on the sequence length. Finally, from this we construct a 1-NN graph by keeping only the closest node to each node and setting all other edges to zero.
\begin{equation}
G(i, j) = \begin{cases} 
            0 & \text{if } W_t(i, j) > \min_{\forall j} W_t(i, j) \\
            1 & \text{otherwise}
            \end{cases}
\label{eqn:Graph G}
\end{equation}
The 1-NN temporal graph $G$ defines an adjacency matrix where each node is connected to its nearest neighbor based on the temporally weighted distances $W_t$. We make the links symmetric by setting $G(j,i)=1$ to encode space-temporal distances and form connected components to clusters conveniently. The connected components of Graph in Equ.{\ref{eqn:Graph G}} automatically partition the data into discovered clusters. We then recursively merge these clusters according to the similarity of their feature averages. Throughout this merging process, we identify multiple clustering results and select the second-largest one as the final segmentation of the video. We denote by $n$ the index set of frames belonging to the $n_{th}$ cluster. Consequently, the set of frames in the $n_{th}$ cluster can be written as:

\begin{equation}
V_{all} = \{V_n | n \in \{1,2, ..., K\} \}
\label{eqn:clusterV}
\end{equation}

We uniformly sample frames from every cluster and combine them as the keyframe sequence $V_{s} = \{v_1,v_2,...,v_{K}\}\in \mathbb{R}^{K \times (L-1) \times D}$ from the segmentations. The segmentation can yield the distribution of events contained in the current video sequence, thereby guiding the fine-grained merging in subsequent steps. 

\subsection{Fine-grained Dynamic Bipartite Merging}\label{subsec:bipartite_merge}
Previous method~\citep{kimImageGridCan2024,xuSlowFastLLaVAStrongTrainingFree2024,FreeVA}  either simply concatenates frames and designs prompts for the VLM or uses pooling method on each image token to construct the video representation. \algname propose a novel method to effectively construct video representations from selected frames.  
After acquiring the segmentations of the frame sequence through the previously described method, we implement a dynamic bipartite token merging approach inspired by ToMe~\citep{bolya2022tome} to minimize the number of visual tokens.  For each individual frame in $v_k$, we sequentially divide the $R_i$ tokens into two non-overlap tokens sets $\mathbb{P}$ with $r_i$ tokens and $\mathbb{Q}$ with $R_i - r_i$ tokens at the $i^{th}$ steps, where initial $R_0 = L-1$(exclude CLS token). We set the $r_i = R_0 - \frac{Z}{K}  $ to dynamically merge the image to preserve more tokens under the fixed visual token length $Z$. To obtain the similarity scores, each visual token is split into $H$ heads along channel dimensions, each with $D/H$ channels. 
The similarity score for each token pair is obtained by averaging the cosine similarity scores over all heads following \eqnref{eqn:scores}
\begin{equation}
a^{p_i q_i} = \frac{1}{H} \left[ \sum_{h=1}^{H} \cos\left( \mathbf{p}_h^{(p_i)}, \mathbf{p}_h^{(q_i)} \right) \right]
\label{eqn:scores}
\end{equation}
where $p_i \in \{1,...,r_i\}$ and $q_i \in \{1,...,(R_i - r_i)\}$ are the indexes of patch feature p in set $\mathbb{P}_i$ and $\mathbb{Q}_i$, respectively. We choose the top-$r_i$ token pairs that have the highest similarity scores and combine the matched tokens through pooling. Finally, the remaining tokens from both sets are joined back together, resulting in $R_i - r_i$ after the $i^{th}$ step. In the end, we efficiently encode the video features from $N \times L \times D$ to $K \times (L-\frac{Z}{K}-1) \times D$ to reduce redundancy in the original visual tokens sequence.

%% file: sec/4_experiment.tex
\section{Experiment}\label{sec:experiment}
To evaluate \algname, we employ over 10 benchmarks that test both structured and open-ended video question answering (VQA) capabilities of the method, specifically using the multiple-choice and GPT-assisted benchmarks.
\begin{table*}[t]
\centering
\setlength{\tabcolsep}{6pt}
\renewcommand{\arraystretch}{1.1}
\resizebox{\linewidth}{!}{
\begin{tabular}{lcccccccc}
\toprule
\textbf{Method} & \textbf{LVLM} &\textbf{Base}& \textbf{Frame} & \textbf{NExTQA} & \textbf{EgoSchema} & \textbf{IntentQA}  & \textbf{VideoMME} & \textbf{MVBench}\\
 & \textbf{Size}&\textbf{Model}& \textbf{Length} & acc. (\%) & acc. (\%) & acc. (\%) & acc. (\%) & acc. (\%)  \\
\midrule
\multicolumn{9}{c}{\textit{Models fine-tuned with video data}} \\
\midrule
SeViLA~\citep{yu2023self}  & 7B & CLIP-L + FlanT5 & 8 & 63.6 & 25.7 & 60.9 & 44.6 & - \\
Video-LLaVA~\citep{linVideoLLaVALearningUnited2023} & 7B & ViT-L + Vicuna & 8 & - & 38.4 & - & 40.4 & 41.0 \\
LLaMA-VID~\citep{li2024llamavid} & 7B & EVA-G + Vicuna & 1fps & - & 38.5 & - & - & 41.9 \\
InternVideo2~\citep{wang2024internvideo2} & 7B  & VideoMAEv2-g+InternVL & 16 & 59.1 & 32.1 & - & 41.6 & -  \\
LLaVA-NeXT-Video~\citep{zhang2024llavanext-video} & 7B & CLIP-L + Vicuna & 32 & - & 43.9 & - & 46.5 & 33.7 \\
VideoLLaMA2~\citep{chengVideoLLaMAAdvancingSpatialTemporal2024} & 7B & CLIP-L+ Mistral-Instruct & 32 & - & 51.7 & - & 46.6 & 54.6 \\
VideoChat2~\citep{liMVBenchComprehensiveMultimodal2024}  & 7B & UMT-L+Vicuna & 16 & 61.7 & - &  59.0 & 39.5 & 51.1 \\
LLaVA-OneVision~\citep{li2024llava} & 7B & SigLIP+Qwen-2 & 32 & 79.4 & 60.1 & - & 58.2 & 56.7 \\
VideoLLaMA2~\citep{chengVideoLLaMAAdvancingSpatialTemporal2024} & 46.7B & CLIP-L+ Mistral-Instruct & 32 & - & 53.3 & - & 47.9 & 53.9 \\
LLaVA-NeXT-Video~\citep{zhang2024llavanext-video} & 32B & CLIP-L + Qwen1.5 & 32 & 77.3 & 60.9 & - & 60.2 & - \\
\midrule
\multicolumn{9}{c}{\textit{Training-free approaches}} \\
\midrule
IG-VLM~\citep{kimImageGridCan2024}  & 7B &  & 6 & 63.1 & 35.8 & 60.1  & 39.8 & 41.3 \\
SlowFast~\citep{xuSlowFastLLaVAStrongTrainingFree2024} & 7B & LLaVA-NeXT-image & 50 & 64.2 & 45.5 & 60.1  & 40.4 & 44.8   \\
\textbf{\algname} & 7B &  & dynamic & \textbf{65.7} & \textbf{48.6} & \textbf{61.6}  & \textbf{41.2} & \textbf{45.2}\\
\midrule
IG-VLM~\citep{kimImageGridCan2024} & 8B &  & 6 & 79.9 & 65.6 & 81.8  & 49.4 & 65.3 \\
SlowFast~\citep{xuSlowFastLLaVAStrongTrainingFree2024} & 8B & InternVL2  & 50 & 77.2 & 60.7 & 79.8  & 48.0 & 57.0 \\
\textbf{\algname} & 8B &  & dynamic  &  \textbf{81.4} & \textbf{67.6} &  \textbf{83.0} &  \textbf{59.3} & \textbf{66.2}\\
\midrule
IG-VLM~\citep{kimImageGridCan2024} & 9B &   &6  & 53.2 & 24.8 & 51.4 & 26.5  & 32.0 \\
SlowFast~\citep{xuSlowFastLLaVAStrongTrainingFree2024} & 9B & Qwen-VL-Chat  & 50 & 48.4 & 17.6 &  45.1 &  22.1 & 31.7 \\
\textbf{\algname} & 9B &   & dynamic &  \textbf{53.4} & \textbf{26.0} &  \textbf{51.7} &   \textbf{27.4} & \textbf{34.3} \\
\midrule
IG-VLM~\citep{kimImageGridCan2024} & 26B &   & 6  & 80.6 & 56.0 & 83.1 & 50.8 & 66.4 \\
SlowFast~\citep{xuSlowFastLLaVAStrongTrainingFree2024} & 26B& InternVL2  & 50  & 79.2 & 54.8 & 82.8 & 49.4 & 61.8\\
\textbf{\algname} & 26B &   & dynamic  &  \textbf{81.1} & \textbf{59.2} &  \textbf{83.6} &   \textbf{53.0} & \textbf{68.1}\\
\midrule
IG-VLM~\citep{kimImageGridCan2024}  & 34B &  & 6 & 70.9 & 53.6 & 65.3  & 52.0 & 48.4\\
SlowFast~\citep{xuSlowFastLLaVAStrongTrainingFree2024} & 34B & LLaVA-NeXT-image & 50 & 71.9 & 55.8 & 66.2  & 53.2 & 51.2\\
\textbf{\algname} & 34B &  & dynamic & \textbf{72.9} & \textbf{56.8} &  \textbf{67.5} &  \textbf{53.4} & \textbf{52.9}\\
\bottomrule

\end{tabular}
}
\caption{
    Structured VQA benchmarks results comparing \algname with SOTA training-free approaches as well as models that have been fine-tuned with additional video data from various LVLM architectures.
}
\label{tab:structured_vqa_results}
\vspace{-0.1in}
\end{table*}
\subsection{Evaluation Benchmarks}
\paragraph{Structured VQA benchmarks.}
For structured VQA, we evaluate \algname on a diverse set of multiple-choice benchmarks, including NExTQA~\citep{xiao2021next}, EgoSchema~\citep{mangalam2023egoschema}, IntentQA~\citep{li2023intentqa}, VideoMME~\citep{fuVideoMMEFirstEverComprehensive2024}, and MVBench~\citep{liMVBenchComprehensiveMultimodal2024}, all of which are designed to quantify model's capability on video understanding by selecting the correct answer among predefined options.
Collectively, these benchmarks provide a comprehensive evaluation of \algname to interpret complex, multimodal data and to select accurate, contextually rich responses across varying levels of task structures. 
Notably, we perform experiments on the VideoMME~\citep{fuVideoMMEFirstEverComprehensive2024} benchmark under the ``w/o subs” configuration, which restricts access to subtitles, thereby isolating the model’s reliance on visual and temporal cues in video understanding.

\paragraph{Open-ended VQA benchmarks.}
We also assess \algname on the open-ended VQA tasks, specifically, we evaluate the zero-shot performance of \algname on MSVD-QA~\citep{msvd}, MSRVTT-QA~\citep{msrvtt}, TGIF-QA~\citep{li2016tgif}, ANet-QA~\citep{yu2019activitynet}, and Video-ChatGPT Generation (VCG)~\citep{maaz2023video} benchmarks. 
These benchmarks require the model to autonomously generate free-form responses, simulating real-world question-answering contexts that demand nuanced understanding of video content.
Specifically, for VCG benchmark, we assess across five key dimensions, including Correctness of Information (CI), Detail Orientation (DO), Contextual Understanding (CU), Temporal Understanding (TU), and Consistency (CO).
Following~\citet{FreeVA}, we utilize \textit{GPT-3.5-Turbo-0125} to ensure fair comparisons with other methods.

\subsection{Experimental Setup}

\paragraph{Input video and model setting.} 
In our approach, we uniformly sample each video to $N = 100$ frames. 
Each frame is resized to match the input dimensions of different visual encoder, which then outputs visual tokens plus a \texttt{[CLS]} token.
Following~\citet{sarfraz2019efficient}, we collect the \texttt{[CLS]} token as the coarse-grained feature for each frame and subsequently partition the sequence into $K$ clusters, using recursive iterations of \textit{hierarchical clustering} (\secref{subsec:hierarchical_clustering}) to group semantically similar frames. 

Due to hardware limitations, we set the visual token sequence length $Z$ to either 3680 or 7200, corresponding to model sizes of 7B and 34B, respectively.
To optimize feature representation and mitigate token redundancy, we dynamically adjust the merge ratio based on the formulation $r = Z/K$. 

\subsection{Main Results}\label{subsec:main_results}
\paragraph{Structured VQA benchmarks.}\label{para:structured_vqa_results} 
As shown in \tabref{tab:structured_vqa_results}, \algname outperforms all the training-free approaches as well as all the fine-tuned models across all benchmarks by clear margins. 
Notably, \algname demonstrated its adaptability to diverse video understanding contexts, achieving superior accuracy even compared to models that employ extensive fine-tuning. 
For example, on NExTQA~\citep{xiao2021next}, \algname sets a new state-of-the-art performance by achieving 81.4\% correctness when pairing with InternVL2-8B base model, which is significantly higher than the 59.1\% accuracy achieved by InternVideo2~\citep{wang2024internvideo2}. 
Similarly, performance on EgoSchema~\citep{mangalam2023egoschema}, IntentQA~\citep{li2023intentqa}, VideoMME~\citep{fuVideoMMEFirstEverComprehensive2024}, and MVBench~\citep{liMVBenchComprehensiveMultimodal2024} illustrate the effectiveness of \algname in handling task-specific reasoning within video content.
%
%
\paragraph{Open-ended VQA benchmarks.}\label{para:unstructured_vqa_results}
\begin{table*}[ht]
\centering
\setlength{\tabcolsep}{2pt}
\renewcommand{\arraystretch}{1.1}
\resizebox{\linewidth}{!}{
\begin{tabular}{lccccccccccccccccc}
\toprule
\textbf{Method} & \textbf{LVLM} & \textbf{Base} & \textbf{Frame} & \textbf{MSVD-QA} & \textbf{MSRVTT-QA} & \textbf{TGIF-QA} & \textbf{ANet-QA} & \multicolumn{6}{c}{\textbf{Video-ChatGPT (VCG Benchmark)}} \\
 & \textbf{Size} & \textbf{Model} & \textbf{Length} & acc./score & acc./score & acc./score & acc./score & \textbf{CI} & \textbf{DO} & \textbf{CU} & \textbf{TU} & \textbf{CO} & \textbf{Average} \\

\midrule
\multicolumn{14}{c}{\textit{Models fine-tuned with video data}} \\
\midrule

Video-ChatGPT\citep{maazVideoChatGPTDetailedVideo2023} & 7B & CLIP-L+Vicuna & 6 & 64.9/3.5 & 49.3/2.9 & 51.4/3.0 & 35.2/2.7 & 2.50 & 2.57 & 2.69 & 2.16 & 2.20 & 2.42 \\
VideoGPT+\citep{MbzuaioryxVideoChatGPT2024} & 3.8B & CLIP-L+Phi-3-mini & 16 & 72.4/3.9 & 60.6/3.6 & 74.6/4.1 & 50.6/3.6 & 3.27 & 3.18 & 3.74 & 2.83 & 3.39 & 3.28 \\
Video-LLava\citep{linVideoLLaVALearningUnited2023} & 7B & ViT-L + Vicuna & 8 & 70.7/3.9 & 59.2/3.5 & 70.0/4.0 & 45.3/3.3 & - & - & - & - & - & - \\
MovieChat\citep{songMovieChatDenseToken2024} & 7B & CLIP-G+Vicuna & 2048 & 75.2/3.8 & 52.7/2.6 & - & 45.7/3.4 & 2.76 & 2.93 & 3.01 & 2.24 & 2.42 & 2.67 \\
LLama-VID\citep{li2024llamavid} & 13B & EVA-G+Vicuna & 1fps & 69.7/3.7 & 57.7/3.2 & - & 47.4/3.3 & 2.96 & 3.00 & 3.53 & 2.46 & 2.51 & 2.89 \\
VideoChat2\citep{liMVBenchComprehensiveMultimodal2024}  & 7B & UMT-L+Vicuna & 16 & 70.0/3.9 & 54.1/3.3 & - & 49.1/3.3 & 3.02 & 2.88 & 3.51 & 2.26 & 2.81 & 2.98 \\
Vista-LLAMA\citep{maVistaLLaMAReliableVideo2023} & 7B & CLIP-L+Vicuna & 16 & 65.3/3.6 & 60.5/3.3 & - & 48.3/3.3 & 2.44 & 2.64 & 3.18 & 2.26 & 2.31 & 2.57 \\
Video-LLama2\citep{chengVideoLLaMAAdvancingSpatialTemporal2024} & 7B & CLIP-L+Mistral-Instruct & 32 & 70.9/3.8 & - & - & 50.2/3.3 & 3.16 & 3.08 & 3.69 & 2.56 & 3.14 & 3.13 \\
PLLaVA\citep{xuPLLaVAParameterfreeLLaVA2024} & 7B & CLIP-L+Vicuna & 16 & 76.6/4.1 & 62.0/3.5 & 77.5/4.1 & 56.3/3.5 & - & - & - & - & - & - \\
LLaVA-NeXT-Video\citep{zhang2024llavanext-video} & 7B & CLIP-L + Vicuna & 32 & - & - & - & 53.5/3.5 & - & - & - & - & - & - \\
Video-LLAMA2\citep{chengVideoLLaMAAdvancingSpatialTemporal2024} & 46.7B & CLIP-L+Mistral-Instruct & 32 & 70.5/3.8 & - & - & 50.3/3.4 & 3.08 & 3.11 & 3.64 & 2.67 & 3.26 & 3.15 \\
LLaVA-NeXT-Video\citep{zhang2024llavanext-video} & 34B & CLIP-L + Qwen1.5 & 32 & - & - & - & 58.8/3.4 & - & - & - & - & - & - \\
\midrule
\multicolumn{14}{c}{\textit{Training-free approaches}} \\
\midrule
IG-VLM-7B\citep{kimImageGridCan2024}  & 7B & LLaVA-NeXT-image & 6 & 78.3/3.9 & 63.7/3.4 & 72.7/4.0 & 53.8/3.2 & 3.11 & 2.78 & 3.48 & 2.44 & 3.29 & 3.03 \\
IG-VLM-34B\citep{kimImageGridCan2024}  & 34B & LLaVA-NeXT-image & 6 & 79.6/4.1 & 62.4/3.5 & 79.1/4.2 & 58.4/3.5 & 3.21 & 2.87 & 3.54 & 2.51 & 3.34 & 3.09 \\
SlowFast-7B\citep{xuSlowFastLLaVAStrongTrainingFree2024}  & 7B & LLaVA-NeXT-image & 50 & 78.7/3.9 & 66.2/3.4 & 77.5/4.0 & 53.9/3.1 & 3.09 & 2.70 & 3.57 & 2.52 & 3.35 & 3.04 \\
SlowFast-34B\citep{xuSlowFastLLaVAStrongTrainingFree2024} & 34B & LLaVA-NeXT-image & 50 & 78.7/4.1 & \textbf{67.1/3.7} & 80.6/\textbf{4.3} & 58.8/3.5 & \textbf{3.48} & \textbf{2.96} & \textbf{3.84} & 2.70 & 3.54 & 3.30 \\
\midrule
\textbf{\algname} & 7B & LLaVA-NeXT-iamge & dynamic & 77.6/3.9 & 64.1/3.4 & 78.0/4.0 & 54.3/3.2 & 3.12 & 2.93 & 3.72 & 2.52 &3.52 & 3.16 \\
\textbf{\algname} & 34B & LLaVA-NeXT-image & dynamic & \textbf{79.6/4.1} & \underline{66.2/3.6} & \textbf{80.7}/\underline{4.2} & \textbf{59.0}/\textbf{3.5} & \underline{3.45} & 2.95 & 3.82 & \underline{\textbf{2.72}} & \underline{\textbf{3.63}} & \underline{\textbf{3.32}} \\
\bottomrule
\end{tabular}
}
\caption{
    Open-ended VQA benchmark results for \algname, demonstrating competitive performance across accuracy and detailed video understanding metrics. 
    Scores indicate the robustness of \algname in generating contextually rich responses without finetuning.
}
\vspace{-0.15in}
\label{tab:open_ended_vqa_results}
\end{table*}

In open-ended settings, as shown in \tabref{tab:open_ended_vqa_results}, \algname also demonstrates competitive zero-shot capabilities. 
It consistently outperforms existing methods on benchmarks including MSVD-QA~\citep{msvd}, MSRVTT-QA~\citep{msrvtt}, TGIF-QA~\citep{li2016tgif}, and ANet-QA~\citep{yu2019activitynet}. 
Notably, \algname performs exceptionally well on Video-ChatGPT~\citep{maaz2023video}, surpassing even those methods that require fine-tuning, highlighting its robustness in zero-shot settings.
\algname’s ability to maintain high accuracy without fine-tuning underscores the strength of its adaptive framework in handling open-ended, real-world VQA tasks, achieving high scores across dimensions like correctness, contextual understanding, and temporal understanding.

%% file: sec/5_analysis.tex
\section{Analysis}\label{sec:Analysis}
\subsection{Scalabity on Base Model Size}\label{subsec:model_size_analysis}
As shown in \tabref{tab:structured_vqa_results}, scaling up model sizes significantly enhances the performance of \algname across structured VQA tasks, demonstrating notable gains over performance with 7B base models.
Specifically, with a 34B model, \algname achieves a 7.2\% accuracy increase on NExTQA~\citep{xiao2021next}, an 8.2\% boost on EgoSchema~\citep{mangalam2023egoschema}, and a 7.7\% improvement on MVBench~\citep{liMVBenchComprehensiveMultimodal2024}. 
These more substantial increments reflect the method’s enhanced capacity to capture and reason over complex spatial-temporal interactions, particularly in tasks requiring nuanced contextual comprehension.

Compared to other training-free approaches at 34B, \algname consistently outperforms IG-VLM~\citep{kimImageGridCan2024} and SlowFast-LLaVA~\citep{xuSlowFastLLaVAStrongTrainingFree2024} across benchmarks, with clear accuracy advantages on tasks like VideoMME~\citep{fuVideoMMEFirstEverComprehensive2024}. 
This suggests that the proposed hierarchical clustering and bipartite token merging mechanisms are particularly effective in leveraging the additional model capacity to retain critical semantic information, even in the absence of fine-tuning.

In the open-ended VQA tasks shown in \tabref{tab:open_ended_vqa_results}, our performance with a 34B model also surpasses its 7B counterpart, reinforcing the scalability of our approach. 
For instance, \algname achieves a 2.3-point average increase across correctness and contextual understanding dimensions of VCG benchmark, as well as improvements in detail orientation and temporal understanding. 
These gains further highlight the robustness of \algname's adaptive framework, which scales efficiently with model size to provide more contextually enriched and temporally accurate responses across various VQA benchmarks.

\subsection{Performance w.r.t. Video Lengths}
\begin{figure}[ht]
    \centering
    \includegraphics[width=.95\linewidth]{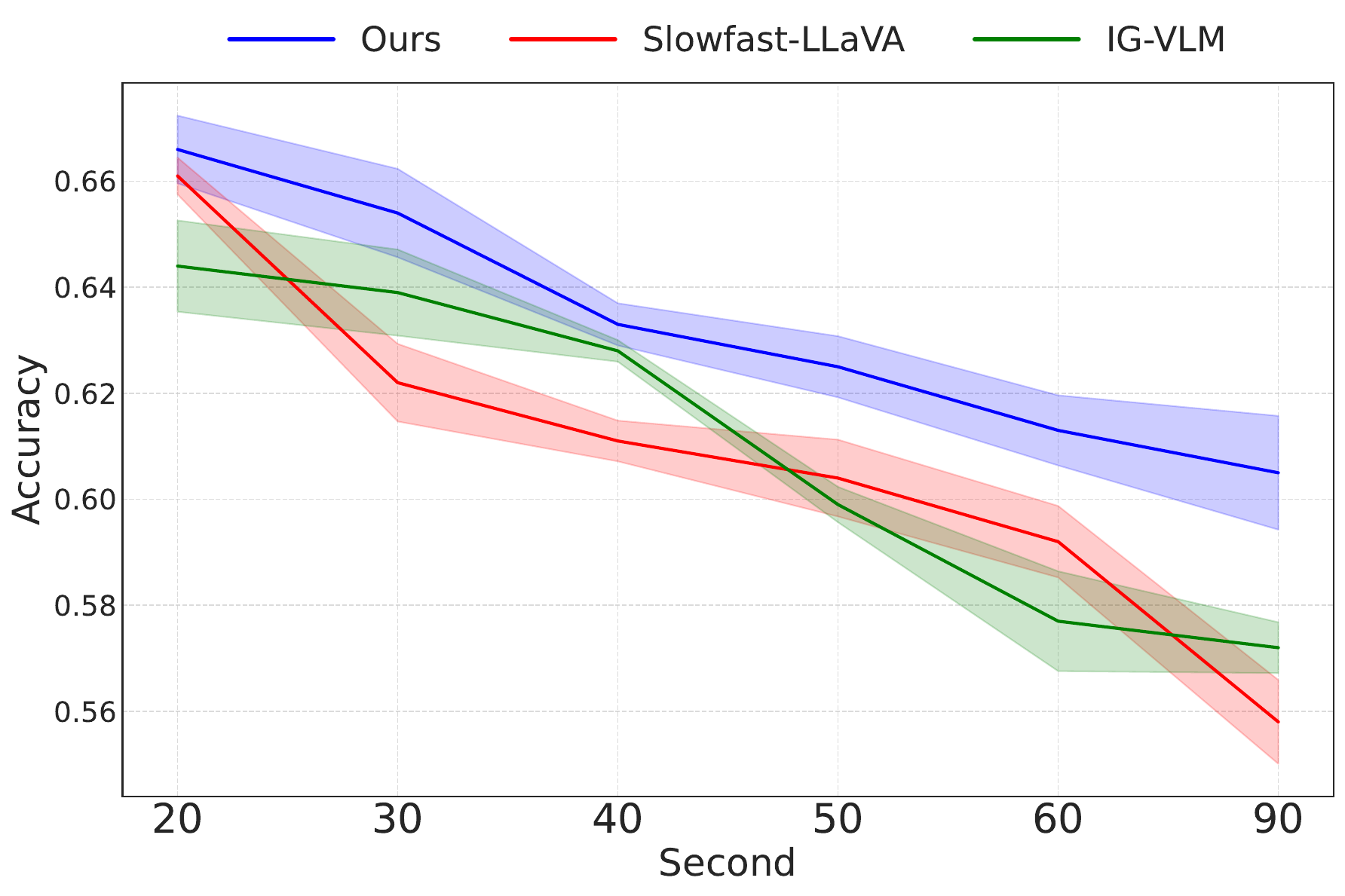} \\
    \includegraphics[width=.95\linewidth]{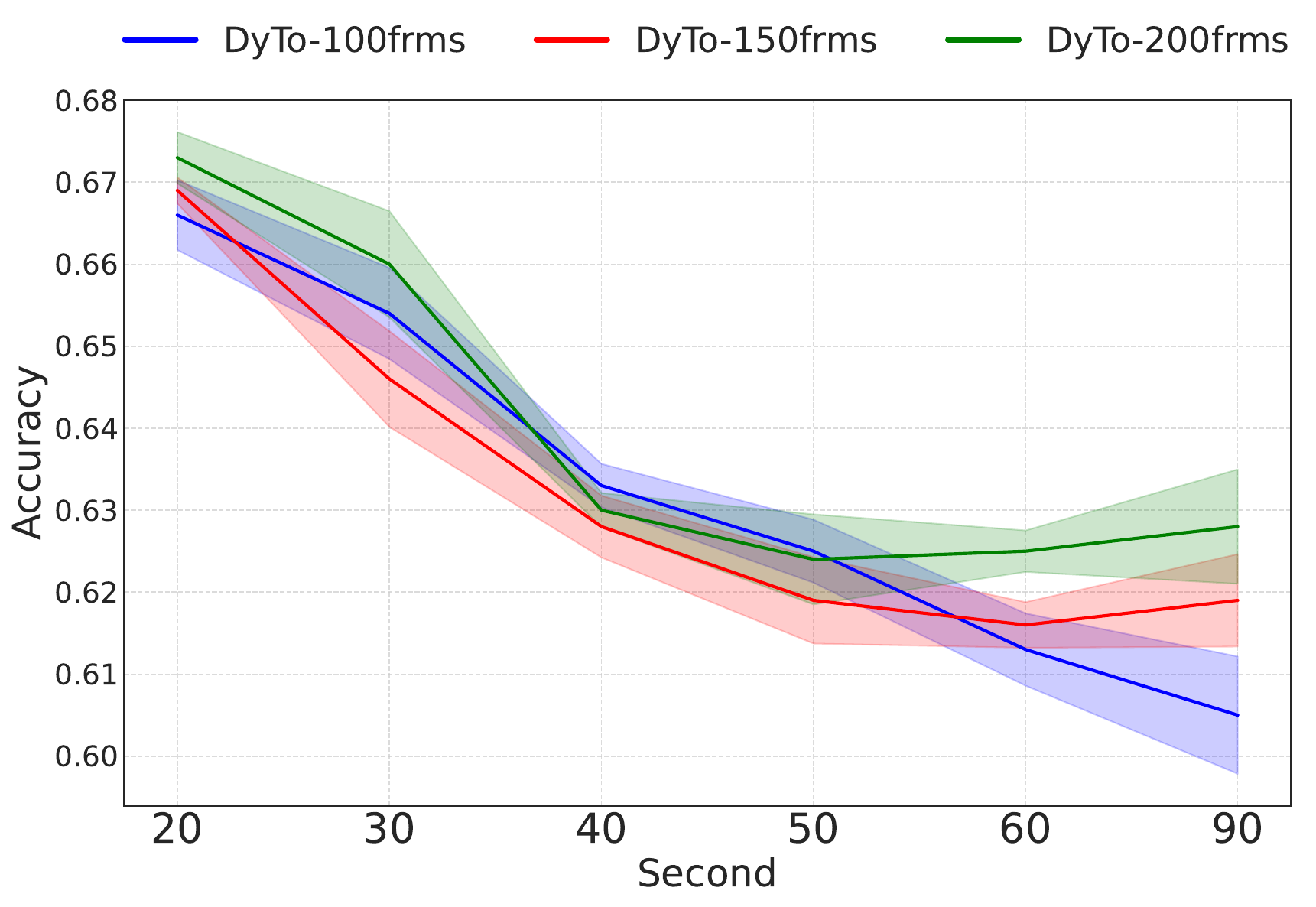}
    \caption{
        \textit{Top}: Performance comparison of baseline method under various video lengths. 
        \textit{Bottom}: Effect of different input sampling lengths under various video lengths.
        %
        %
        %
    }
    \label{fig:video_lengths_ablation}
    \vspace{-0.18in}
\end{figure}
%

%
We analyze the performance of \algname across varying video lengths to better demonstrate its robustness, particularly in maintaining accuracy with longer videos. 
As shown in \figref{fig:video_lengths_ablation}~(\textit{Top}), while the performance of \algname does show some degradation as video length increases, the decline is significantly less pronounced compared to other methods like IG-VLM~\citep{kimImageGridCan2024} and SlowFast-LLaVA~\citep{xuSlowFastLLaVAStrongTrainingFree2024}. 
This stability is largely attributed to \algname's adaptive, video-dependent dynamic token merging, which effectively prioritizes essential frames and contextual tokens, preserving key information even as the video sequence lengthens.

In \figref{fig:video_lengths_ablation}~(\textit{Bottom}), we further examine the impact of increasing the number of sampled frames on performance with extended videos. 
%
As the number of sampled frames increases, \algname's accuracy on longer videos notably improves, demonstrating its capacity to capture detailed temporal and contextual information more effectively than competing methods. 
This improvement highlights the advantage of \algname's hierarchical clustering and bipartite token merging, which dynamically adjust to sample frames that represent critical moments, ensuring more comprehensive and accurate video understanding.
%

%
This indicates a robust generalization across varying video lengths, reinforcing \algname’s strength in managing diverse content scales without substantial accuracy loss or the need for fine-tuning.

\subsection{Ablation of Clustering and Merging Module}\label{subsec:ablation study}
The experimental results from the table demonstrate significant differences in performance when \algname adopts different strategies.verall, DyTo consistently outperforms both the baseline (SlowFast) and its ablated variants. Notably, on NExTQA, DyTo w/o Clustering slightly surpasses the full DyTo model. A closer inspection reveals that, due to the relatively short clips in NExTQA, the benefits of clustering become less pronounced, thus minimizing DyTo’s advantage.
Nevertheless, in all other datasets, DyTo exhibits clear improvements over the baseline and ablation models, highlighting its stronger capability in handling more complex or longer video tasks. Furthermore, this performance edge persists regardless of the language model size (7B or 34B), demonstrating DyTo’s effectiveness and robustness in diverse settings.

\subsection{Visualizations of Hierarchical Clustering}\label{subsec:clustering_analysis}
\begin{figure*}[ht]
    \centering
    \includegraphics[width=.9\linewidth]{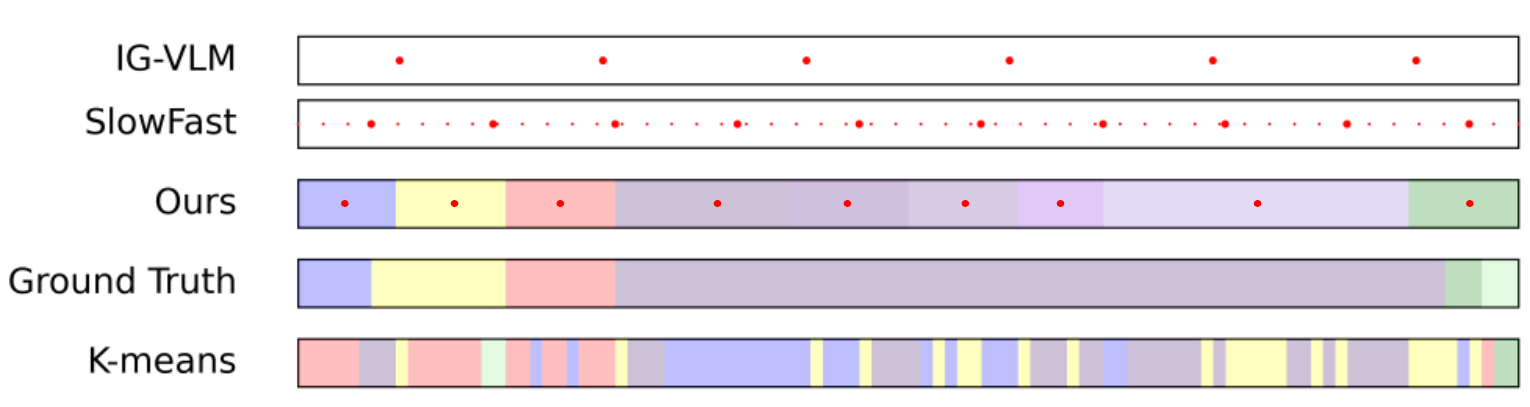}
    \caption{
        The sampling method and clustering module output visualization on a video.
        Our method offers more comprehensive video representation frames compared to other methods.
        %
    }
    \label{fig:classification}
\end{figure*}

\begin{figure*}[ht]
    \centering
    \includegraphics[width=.9\linewidth]{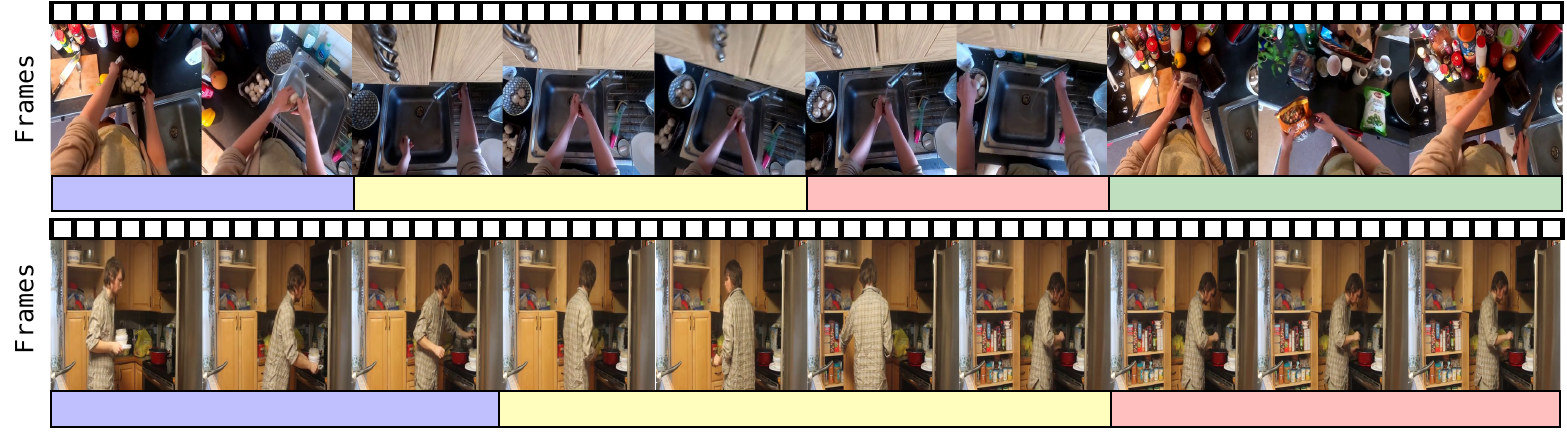}
    \vspace{-0.05in}
    \caption{
        Clustering module output example from videos. 
        Colors indicate different events in temporal order. The differences are clearly visible in the video clips.
        %
    }
    \label{fig:classification-vis}
    \vspace{-0.15in}
\end{figure*}

To further illustrate \algname's improved sampling and segmentation, \figref{fig:classification} visualizes the clustering output from our method alongside other approaches, including IG-VLM~\citep{kimImageGridCan2024} and SlowFast-LLaVA~\citep{xuSlowFastLLaVAStrongTrainingFree2024}.
By mapping the video events using Hungarian matching, we demonstrate that \algname has a clear advantage in accurately capturing and segmenting all critical events without omissions.
Unlike IG-VLM and SlowFast-LLaVA, which may overlook certain events by failing to sample any key-frame from them--such as IG-VLM missing the first event of the video completely--\algname reliably identifies each segment's temporal boundaries, allowing for precise key-frame selection.
\begin{table}[t]
\centering
\setlength{\tabcolsep}{1pt}
\renewcommand{\arraystretch}{1.1}
\resizebox{\linewidth}{!}{
\begin{tabular}{lcccccc}
\toprule
\textbf{Method}  &\textbf{LLM}& \textbf{NExTQA} & \textbf{EgoSchema} & \textbf{IntentQA}  & \textbf{VideoMME} &\textbf{MVBench}\\
& \textbf{Size} & acc. & acc. & acc. & acc. & acc. \\
\midrule
SlowFast & 7B & 64.2 & 45.5 & 60.1  & 40.4 & 44.8   \\
\algname w/o Clustering  & 7B  & 65.6 & 47.8 & 61.4 & 42.3 & 44.8 \\
\algname w/o Token Merge & 7B  & 64.9 & 45.6 & 60.5 & 41.2 & 44.6 \\
\algname  & 7B  & \textbf{65.7} & \textbf{48.6} & \textbf{61.6} & \textbf{42.7} & \textbf{45.2} \\

\midrule
SlowFast & 34B & 71.9 & 55.8 & 66.2  & 53.2 & 51.2\\
\algname w/o Clustering  & 34B  & \textbf{73.2} & 55.8 & 66.8 & 52.5 & 52.7 \\
\algname w/o Token Merge & 34B  & 72.2 & 56.0 & 66.3 & 51.7 & 51.8 \\
\algname  & 34B  & 72.9 & \textbf{56.8} & \textbf{67.3} & \textbf{53.4} & \textbf{52.9} \\

\bottomrule
\end{tabular}
}
\caption{
    Ablation Study on clustering and token merging modules with parameter sizes ranging from 7B to 34B.
}
\vspace{-0.15in}
\label{tab:structured_vqa_results_34b}
\end{table}
Specifically, although SlowFast-LLaVA aims to mitigate this limitation by sampling 50 frames with coarse $4 \times 4$ tokens as the fast part input, shown as smaller red dots in \figref{fig:classification}, its main performance heavily depends on the 10 frames (the slow part) that carry the main temporal load~\citep{xuSlowFastLLaVAStrongTrainingFree2024}.
Furthermore, \figref{fig:classification-vis}, where each color represents different events segmented over time, highlights how \algname's clustering naturally organizes video content in a temporally coherent manner.
This helps \algname to achieve a higher fidelity in key-frame selection, making it better suited for scenarios requiring comprehensive, action-specific video understanding.
%

%

\begin{figure}[ht]
    \centering
    \includegraphics[width=.95\linewidth]{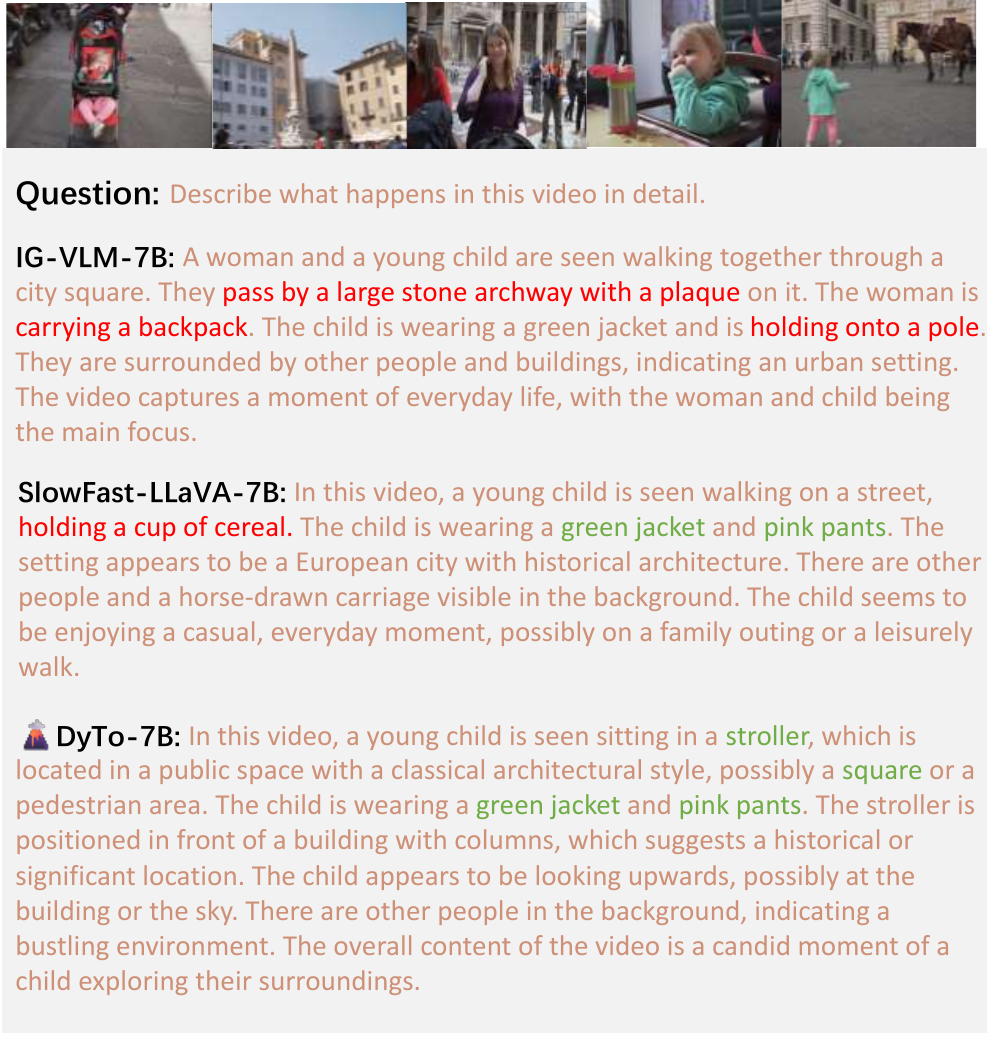}
    \vspace{-0.1in}
    \caption{
        Example from NExTQA benchmark. \textcolor{red}{red} part denotes the incorrect or hallucination content.  \textcolor{green}{green} part denotes the correct object content. \algname demonstrates significantly better performance compared to the other two methods.
    }
    \label{fig:qualitative_example}
    \vspace{-0.25in}
\end{figure}
\subsection{Qualitative Case Study}
\figref{fig:qualitative_example} provides a qualitative illustration of \algname's robust performance in a complex zero-shot video understanding scenario from the NExTQA benchmark. Based models is LLaVA-NeXT-Image-7B. 
In this instance, \algname accurately identifies and preserves critical visual details, such as the red toy on the stroller, the child's attire, and the architectural background.   
%
Due to the simple uniform sampling and pooling method, IG-VLM and SlowFast-LLaVA both output the hallucination content (\textcolor{red}{carrying a backpack} and \textcolor{red}{holding a cup of cereal} respectively). Unlike these methods, \algname captures nuanced visual cues and contextual elements. 
\algname's hierarchical clustering, which segments video frames dynamically, alongside its bipartite merging strategy that avoids excessive compression that might lead to semantic loss, result in a more detailed and accurate scene understanding. 

%% file: sec/6_conclusion.tex
\section{Conclusion and Future Work}\label{sec:conclusion}
In this paper, we introduced \algname, a training-free framework designed for dynamic token merging in zero-shot video understanding. 
Through a novel combination of hierarchical frame selection and bipartite token merging, \algname addresses the challenges of maintaining semantic richness while enhancing computational efficiency. 
Experimental results demonstrate that \algname achieves state-of-the-art performance across multiple structured and open-ended VQA benchmarks, outperforming both fine-tuned and other training-free models. 
By dynamically adjusting token granularity based on frame content, \algname successfully captures critical spatial-temporal details, offering a scalable solution that adapts to varying video lengths and complexities. 
\algname not only sets a new standard in zero-shot video tasks but also paves the way for more efficient and contextually aware video understanding. 
Future work may explore extending \algname to enhancing token adaptability for real-time applications, further pushing the boundaries of training-free video comprehension.

%% file: sec/X_suppl.tex
\appendix
\setcounter{page}{1}
\onecolumn
\section{Time Consumption Experiment}\label{app:time_consumption}
we conducted the experiments using the same hardware specifications. The table below shows the time consumption for inference with 500 samples from EgoSchema using a single NVIDIA A100 GPU.
\begin{table}[ht]
    \centering
    \setlength{\tabcolsep}{8pt}
    \renewcommand{\arraystretch}{1}
    \begin{tabular}{c|cc}
        \midrule
                & SlowFast-LLaVA & \algname   \\
        \midrule
        Dataset  & \multicolumn{2}{c}{Egoschema } \\
        \midrule
        Model & \multicolumn{2}{c}{LLaVA-NeXT-34B}  \\ 
        \midrule
        Input  &  $50 + 10$ frames&   100 frames \\
        \midrule
        Merge Strategy & Pooling & Dynamic token merging \\ 
        \midrule
        Device & \multicolumn{2}{c}{1 Nvidia A100 GPU} \\
        \midrule
        Time Consumption & 5.74 s/item & 6.22 s/item \\
        \midrule
    \end{tabular}
    \vspace{-0.1in}
    \caption{time consumption}
    \label{tab:time_consumption}
\end{table}

As shown in the table, the difference in time consumption is negligible. Although our method is slightly slower than SlowFast, we think it may be attributed to hardware optimizations or variance.

\section{Implementation details} 
All evaluations could be conducted on a single Nvidia A100 80G graphics card. 
To accelerate inference, we use a Linux server equipped with 8 Nvidia A100 80G cards. 
%
%
We carry out our evaluation across three model series and five model size. The  weights for these models are available on Huggingface\footnote{https://huggingface.co/collections/liuhaotian/llava-16-65b9e40155f60fd046a5ccf2}\footnote{https://huggingface.co/OpenGVLab/InternVL2}\footnote{https://huggingface.co/Qwen/Qwen-VL-Chat}.
We implement rotary position embedding (RoPE) and apply a scaling factor of 2, extending the original context length from 4096 to 8192 tokens.

\clearpage

\section{Visualizations of Dynamic Bipartite Merging}

\begin{figure*}[htbp]
    \centering
    \includegraphics[width=\linewidth]{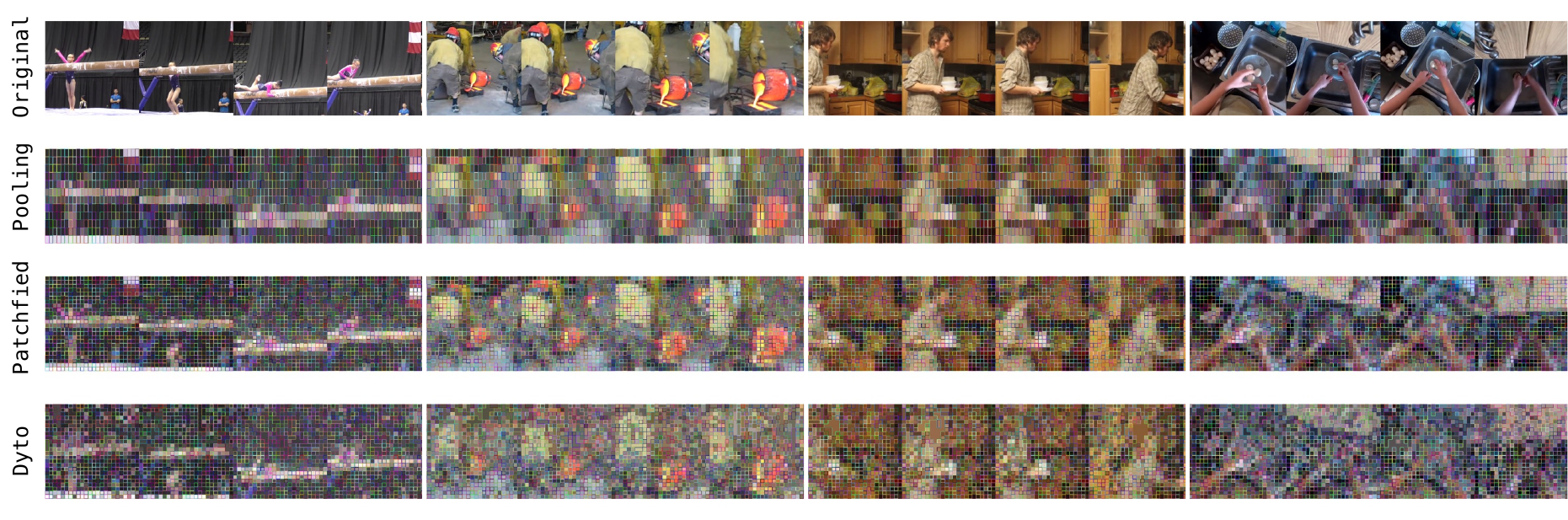}
    \caption{
        Results of merging on video frames.From top to bottom is original images,after pooling, patchfied, and the proposed bipartite merging. Patchfied means 
    }
    \label{fig:tomevis}
\end{figure*}

To help understand dynamic token merging effectively, we provide the visualization comparing our method with the pooling method. As shown in \figref{fig:tomevis}, the proposed approach effectively maintains the object's actions while making every effort to prevent the disruption of the original spatial information. We set the constant merge ratio of r=288 to enable a convenient and fair comparison with the pooling method, while r is a dynamic integer that varies based on the number of clusters in \algname. It is important to emphasize that our proposed token merging method operates without the need for any labels.
To create the visualizations in Figure \ref{fig:tomevis} , we follow each final merged token back to its original
input patches. For each token, we color its corresponding input patches, referred to as "Patchified," using the average color of that region. To ensure that different tokens are distinguishable, we assign each token a random border color. It's important to note that tokens do not necessarily correspond to contiguous input regions. The only spatial information comes from the position encodings.
\clearpage

\section{More Qualitative Cases}\label{app:time_consumption}

\begin{figure*}[ht]
    \centering
    \includegraphics[width=.9\linewidth]{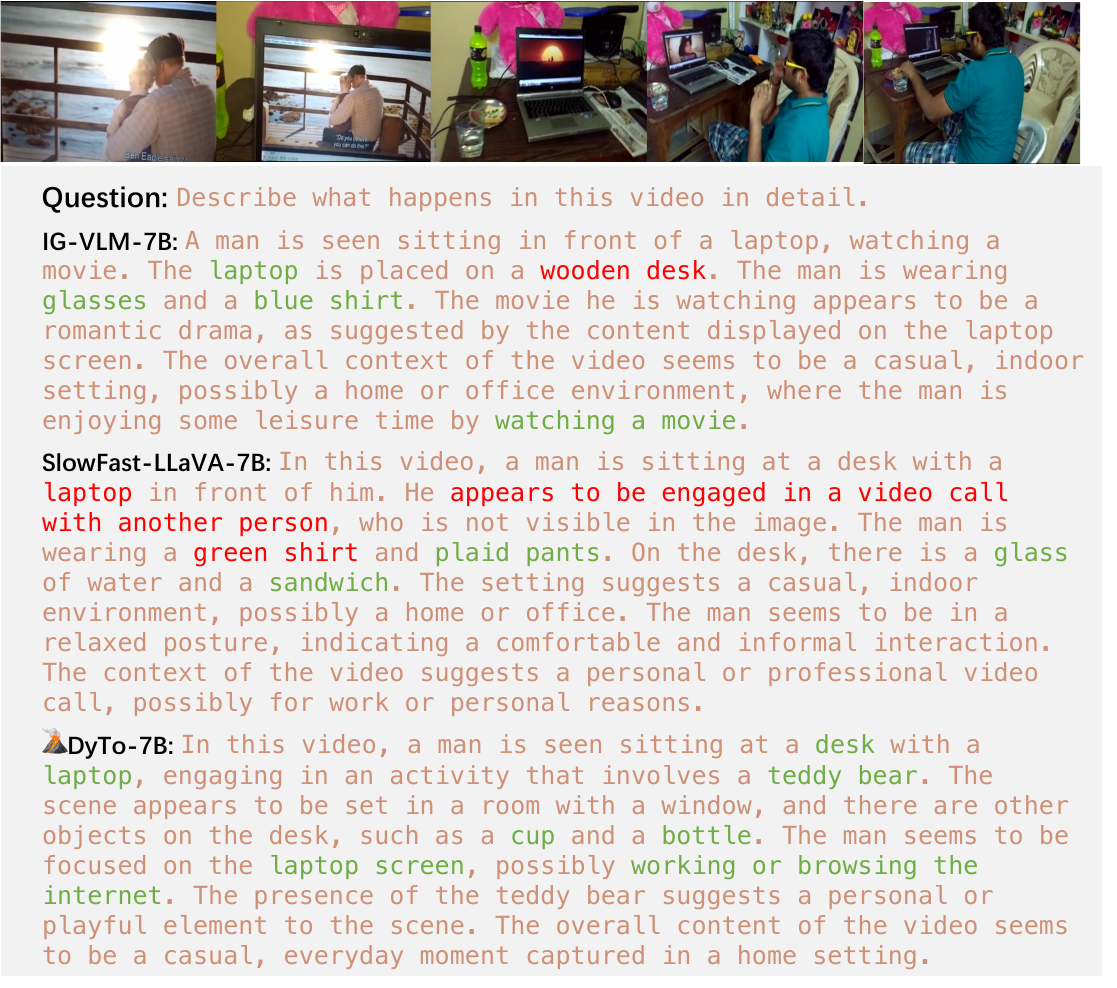}
    \caption{
        Example from MVBench benchmark. \textcolor{red}{red} part denotes the incorrect or hallucination content.  \textcolor{green}{green} part denotes the correct object content. 
    }
    \label{fig:qualitative_example}
\end{figure*}

\clearpage
\begin{figure*}[ht]
    \centering
    \includegraphics[width=.9\linewidth]{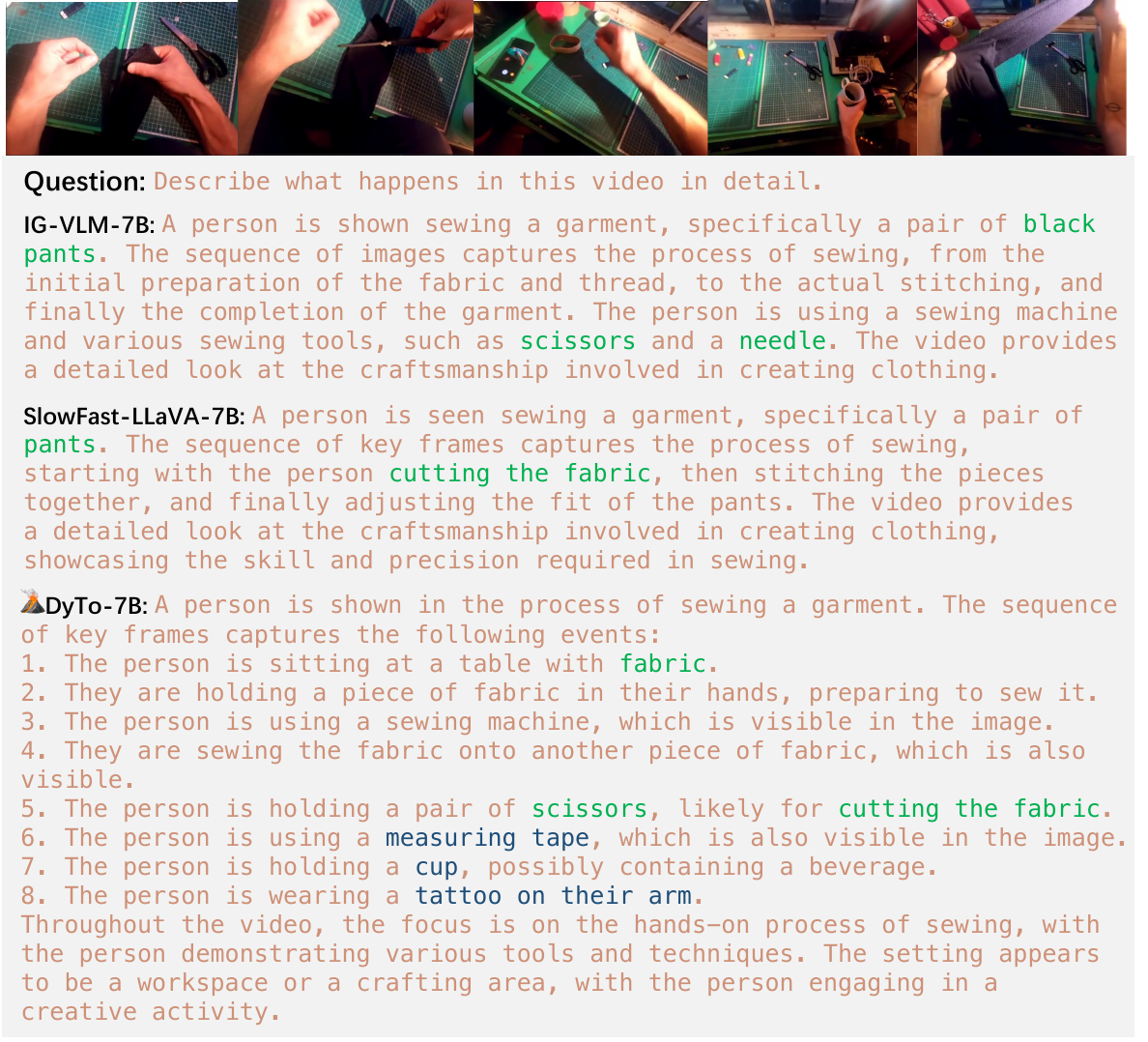}
    \caption{
        Example from Egoschema benchmark. \textcolor{green}{green} part denotes the correct object content. \textcolor{blue}{blue} part denotes the missing content in IG-VLM and SlowFast-LLaVA answers. \algname can effectively capture the events in the video.
    }
    \label{fig:qualitative_example}
\end{figure*}